\documentclass[runningheads]{llncs}
\usepackage[T1]{fontenc}
\usepackage{graphicx}
\usepackage{hyperref}
\usepackage{multirow}%
\usepackage{amsmath,amssymb,amsfonts}%
\usepackage{mathrsfs}%
\usepackage[title]{appendix}%
\usepackage{xcolor}%
\usepackage{textcomp}%
\usepackage{manyfoot}%
\usepackage{booktabs}%
\usepackage{algorithm}%
\usepackage{algorithmicx}%
\usepackage{algpseudocode}%
\usepackage{listings}%
\usepackage[T1]{fontenc}
\usepackage{mathptmx}
\usepackage{hyperref}

\begin{document}
\title{Mitigating Hallucinations in Zero-Shot Scientific Summarisation: A Pilot Study}
%
\author{Imane Jaaouine, Ross D. King}
%
%
\institute{Department of Chemical Engineering and Biotechnology \\ University of Cambridge}
\maketitle              
\begin{abstract}
Large language models (LLMs) produce context inconsistency hallucinations, which are LLM-generated outputs that are misaligned with the user prompt. This research project investigates whether prompt engineering (PE) methods can mitigate context inconsistency hallucinations in zero-shot LLM summarisation of scientific texts, where zero-shot indicates that the LLM relies purely on its pre-training data. Across eight yeast biotechnology research paper abstracts, six instruction-tuned LLMs were prompted with seven methods: a baseline prompt, two levels of increasing instruction complexity (PE-1 and PE-2), two levels of context repetition (CR-K1 and CR-K2), and two levels of random addition (RA-K1 and RA-K2). Context repetition involved the identification and repetition of $K$ key sentences from the abstract, whereas random addition involved the repetition of $K$ randomly selected sentences from the abstract, where $K\!\in\!\{1,2\}$. A total of 336 LLM-generated summaries were evaluated using six metrics: ROUGE-1, ROUGE-2, ROUGE-L, BERTScore, METEOR, and cosine similarity, which were used to compute the lexical and semantic alignment between the summaries and the abstracts. Four hypotheses on the effects of prompt methods on summary alignment with the reference text were tested. Statistical analysis on 3744 collected datapoints was performed using bias-corrected and accelerated (BCa) bootstrap confidence intervals and Wilcoxon signed-rank tests with Bonferroni-Holm correction. The results demonstrated that CR and RA significantly improve the lexical alignment of LLM-generated summaries with the abstracts. These findings indicate that prompt engineering has the potential to impact hallucinations in zero-shot scientific summarisation tasks.


\keywords{Large Language Models \and Prompt Engineering \and Scientific Summarisation \and Information Retrieval}
\vspace{0.1cm} \\

\end{abstract}

\newpage
\section{Introduction}

Large language models (LLMs) have gained global public attention, increasing in popularity across almost all fields \cite{llmsnature}, including academia \cite{surveyllms}. However, LLMs confidently produce hallucinations, defined as LLM-generated content that is factually incorrect, nonsensical or misaligned with the user prompt \cite{Huang_2024}. Hallucinations, through their distortion of knowledge, pose a risk to the integrity of science \cite{protectscience}, particularly when considering LLM-generated scientific knowledge. The trustworthiness of LLMs is a major concern due to the persuasive tone in which LLMs present incorrect information to users \cite{Huang_2024}, such as false citations \cite{llmsnature}, and the difficulty in detecting hallucinations 
\cite{Huang_2024}. Hallucinations are concerning when considering real-world information retrieval systems, such as chatbots and search engines \cite{Huang_2024}, where the distribution of incorrect LLM-generated scientific information can occur, particularly in challenging text generation tasks such as automatic summarisation \cite{elementawaresummmarisation}. 

As the number of scientific publications increases by millions every year \cite{AuerOelenHarisStockerDSouzaFarfarVogtPrinzWiensJaradeh+2020+516+529,scientificsummarisation}, it is becoming more difficult for researchers to explore scientific literature \cite{scientificsummarisation}. LLMs are increasingly used in academia \cite{surveyllms}, with novel scientific publications outside of LLMs' knowledge boundaries \cite{Huang_2024}. Therefore, it is critical to investigate the reduction of hallucinations within LLM-generated scientific information. To address this challenge, prompt engineering can be applied. Prompt engineering modifies LLM inputs, known as prompts, and has been shown to improve LLM performance \cite{jm3,sahoo2025systematicsurveypromptengineering}. Prior research by Kawamura et al. (2024) \cite{financialsumm} investigated the impact of prompt design on LLM summarisation within the domain of financial news, where \texttt{Meta-Llama-3-8B-Instruct} was prompted to generate a title based on the body of a financial news article. Within the experiment, four prompts of varying instruction complexity were tested across zero-shot generation, where the LLM relies on its existing training data, and fine-tuned generation, where additional training is provided. The LLM-generated summaries, in the form of financial news titles, were evaluated using ROUGE-1. Wang et al. (2023) \cite{elementawaresummmarisation} explored LLM summarisation of news summary datasets, using ROUGE-1, ROUGE-2, ROUGE-L, and BERTScore to evaluate the summaries. Basyal et al. (2023) \cite{basyal2023textsummarizationusinglarge} also tested LLM summarisation using news article datasets, using evaluation metrics such as ROUGE and BERTScore. Furthermore, Shaier et al. (2025) \cite{shaier2025askingagainexploringllm} investigated the effect of repeated questions in prompts within LLM reading comprehension tasks. 

This study aims to build upon existing research by investigating increasing instruction complexity and the repetition of both semantically relevant and randomly selected technical information within scientific summarisation tasks. Overall, seven prompt methods are evaluated: a simple baseline prompt, two prompts of increasing instruction complexity (PE-1 and PE-2), and prompts integrating repeated sentences extracted from the research paper abstracts (CR-K1, CR-K2, RA-K1 and RA-K2). The impact of these prompt methods on context inconsistency hallucinations is measured by proxy through the lexical and semantic alignment between the LLM-generated summaries and the reference text. The hypotheses tested, presented in Section 1.1, focus on whether instruction complexity and repetition can reduce context inconsistency hallucinations within scientific summarisation tasks.

\subsection{Hypotheses}
The hypotheses tested focus on whether instruction complexity and repetition can reduce context inconsistency hallucinations within scientific summarisation tasks.

\vspace*{-0.4cm}

\subsubsection{H1:} Repeating semantically key sentences (CR-K1, CR-K2) in prompts improves lexical alignment between the LLM-generated summaries and the original abstract, compared to the baseline prompt.

\vspace*{-0.4cm}

\subsubsection{H2:} Repeating randomly selected sentences (RA-K1, RA-K2) in prompts improves lexical alignment between the LLM-generated summaries and the original abstract, compared to the baseline prompt.

\vspace*{-0.4cm}

\subsubsection{H3:} Increasing prompt instruction complexity (PE-1, PE-2) improves semantic alignment between the LLM-generated summaries and the original abstract, compared to the baseline prompt.

\vspace*{-0.4cm}

\subsubsection{H4:} Repeating key sentences within the prompt increases alignment with the key sentences when used as the evaluation reference, compared to random addition.

\section{Literature Review}

\subsection{Hallucinations in LLMs}
\subsubsection{Hallucination Classification}

Hallucinations are defined as outputs that are factually wrong or misaligned with the provided context or instructions \cite{yu2024mechanisticunderstandingmitigationlanguage}. Huang et al. (2024) \cite{Huang_2024} proposed a novel system for classifying hallucinations. This system organises hallucinations initially by two dimensions: factuality and faithfulness. Factuality hallucinations are outputs that misrepresent verifiable real-world knowledge or invent information that cannot be checked. Faithfulness hallucinations are outputs that differ from the user's instruction or input context, such as providing irrelevant or logically inconsistent responses. The hallucination type within this study is classified as unfaithful rather than non-factual, as even though misalignment with the inserted paper abstract would constitute non-factual scientific information, the main aim of this study is to improve the alignment of LLM-generated summaries with the provided user input, which in this case are abstracts. This is opposed to aligning with known world knowledge, which is not the case within this study, as the paper abstracts are assumed to be non-parametric and therefore do not constitute part of the LLM's known world knowledge.

Faithfulness hallucinations can be categorised into three subtypes: instruction inconsistency, context inconsistency, and logical inconsistency \cite{Huang_2024}. Within the faithfulness hallucinations classified by Huang et al. (2024) \cite{Huang_2024}, this paper focuses on the reduction of context inconsistency hallucinations, where context refers to the abstracts. The mitigation of these hallucinations is explored through the application of context repetition, random addition and the increase of instructions within prompts.


\subsubsection{Causes of Hallucinations}
Causes of hallucinations in LLMs are multifaceted. Factors related to data, such as misinformation within pre-training corpora and intrinsic knowledge limits, restrict the reliability of the model's output. This often increases the risk of hallucination, especially if the models are pushed outside their knowledge base and pre-defined capability boundaries to respond to queries \cite{Huang_2024}. Within this case study, context inconsistency hallucinations are expected as the paper abstracts that act as the source material for LLM summarisation are assumed to be outside of the models' knowledge bases, and so the chance of hallucination is heightened. 


\paragraph{Training Data Limitations}
The core of LLM functionality lies in their training data \cite{zhou2023limaalignment}, which consists of pre-training data, where LLMs source their knowledge and capabilities, and alignment data, where LLMs learn to align with user instruction and preference \cite{Huang_2024}. The quality of the training data directly impacts the accuracy of model outputs \cite{zhou2023limaalignment,Huang_2024}, and therefore affects the chance of hallucination. There are three types of hallucinations which arise from training data: misinformation and biases, knowledge boundary, and inferior alignment data \cite{Huang_2024}. 




\paragraph{Knowledge Boundaries}
Hallucinations are more likely when the user prompt falls outside of the knowledge boundaries \cite{Huang_2024}. Knowledge boundaries exist within LLMs as pre-training data does not include up-to-date world knowledge or proprietary information. Copyright-sensitive knowledge boundaries exist due to licensing restrictions and copyright laws. This type of knowledge boundary limits the data available for LLMs, particularly in recent scientific research and copyrighted works. Hallucinations arise when LLMs try to generate responses when prompted for copyrighted or up-to-date knowledge \cite{Huang_2024}. Moreover, the static nature of LLM datasets limits their responses to knowledge that existed at the time of their training \cite{zhou2023limaalignment}. This temporal knowledge boundary causes models to generate outdated or hallucinated responses to prompts on recent real-world knowledge, especially in domains such as scientific research \cite{Huang_2024}, where corrections and advancements are continuously made. For this reason, the yeast biotechnology paper abstracts were integrated into the prompt to allow for LLM summarisation, as the content was assumed to be beyond the knowledge boundaries of the selected LLMs.


\subsection{Retrieval-Augmented Generation (RAG)}

Retrieval-Augmented Generation (RAG) integrates external knowledge retrieval with text generation, grounding outputs in verifiable information. Unlike traditional parametric models, RAG relies both on parametric memory stored in the model's weights and on non-parametric memory sources from external knowledge bases such as Wikipedia. By explicitly conditioning outputs on retrieved evidence, RAG reduces reliance on the probabilistic token generation that often leads to hallucinations \cite{lewis2021retrievalaugmentedgenerationknowledgeintensivenlp}.

Recent extensions to RAG have been focused on increasing its domain-specific adaptability. Zhao et al. (2024) \cite{zhao2024retrievalaugmentedgenerationaigeneratedcontent} classified three main approaches: query-based, latent representation-based, and logit-based RAG. The most popular, query-based RAG, involves integrating retrieved documents directly into the input, hence allowing the LLM to process the query with supportive evidence \cite{zhao2024retrievalaugmentedgenerationaigeneratedcontent}. This study investigates methods to improve accuracy within query-based RAG, where paper abstracts are integrated directly into the LLM prompt.

\subsection{Variants of Retrieval-augmented Generation (RAG)}

SummRAG, introduced by Lui et al. (2024) \cite{liu2024robustretrievalbasedsummarization}, adapts the Self-RAG system to summarisation tasks. It integrates the top-k external documents based on semantic similarity and their evaluated relevance to the user prompt \cite{liu2024robustretrievalbasedsummarization}. The work by Kukreja et al. (2024) \cite{10603291} partitions a large dataset into documents, referred to as chunks, and encodes them into vector embeddings (numerical representations). User prompts are also converted into vector embeddings. The top k documents with embedding vectors closest to the user prompt embedding vectors are retrieved as context for the LLM. Cosine similarity is used as one of the distance metrics within the semantic search \cite{10603291}. The method used by Lui et al. (2024) \cite{liu2024datasetslargelanguagemodels} and Kukreja et al. (2024) \cite{10603291} is applied in this study for the identification of top-k sentences within each paper abstract for the context repetition prompt method.

\subsection{Prompt Engineering (PE)}
Instructions given to language models are called prompts. Prompt engineering  (PE) is a practical and cost-effective mitigation method to guide LLM behaviour without changing underlying architecture or parameters. LLMs decode the next token by finding the highest token probability using the prompt as context, $P(w_i|w_{<i})$, where $w_i$ is the input token, $w_{<i}$ are the previous tokens, and $w_{i+1}$ will be the next predicted token. The structure of input prompts can influence the accuracy and factuality of model outputs. Structured prompts explicitly guide the model to produce relevant and grounded responses by including contextual instructions or domain-specific constraints \cite{jm3}.

\section{Experimental Section}

\subsection{Overview}
This research project involved designing, creating, and applying a systematic framework to investigate and evaluate the mitigation of context inconsistency hallucinations within zero-shot summarisation of yeast 
biotechnology abstracts. Each abstract was summarised by six selected instruction-tuned LLMs under seven prompt conditions (baseline, PE-1, PE-2, CR-K1, CR-K2, RA-K1, RA-K2). The corpus consisted of eight open-access yeast-biotechnology papers, listed in Table \ref{tab:papers}, resulting in: 
\begin{equation}\notag
8\;\text{papers}\;\times 6\;\text{LLMs}\;\times 7\;\text{prompts}=336 \text{ LLM-generated summaries}
\end{equation}

Zero-shot prompting is where LLMs are given instructions without being provided with examples of the task \cite{zero-shot}. Within this study, zero-shot prompting indicates that the LLMs were not fine-tuned for scientific summarisation or shown training examples of summaries. This ensures the evaluation of LLM summarisation based only on the LLM's pre-training data and the prompt method.

\subsection{Source Material Selection}
Chen et al. (2025) \cite{tokenlimits} found performance improvement through fine-tuning to be unsuccessful within text summarisation tasks. This was due to the input context exceeding the allowed input tokens for the model, which was 4096 tokens \cite{tokenlimits}. Due to the input token limits, the integration of full papers as LLM-summarisation source material would have been unsuccessful, as the full paper text would only be partially integrated into the prompt before being truncated once the input token limit was reached. Consequently, it was decided that the abstracts would be integrated as the source material for summarisation tasks, as they could be fully integrated into the prompts without exceeding the input token limits, while still providing scientific information for summarisation.

\subsection{Large Language Model Selection}
Six instruction-tuned LLMs, presented in Table \ref{tab:llmssummary} and representing a wide range of parameter sizes and developers, were selected to conduct the scientific summarisation tasks. All the inference calls to LLMs were executed through the Hugging Face Inference API and accessed under various licenses that allowed their usage within non-commercial academic research. The selected LLMs were not changed or redistributed. Additionally, this research follows Hugging Face's Terms of Service and Content Policy. 

\begin{table}[h]
    \centering
    \caption{\textbf{A Summary of the Large Language Models Used.} The developer indicates the company that developed the model, the parameters indicate the model size and complexity, and the license shows the type of license under which each model is released. The selected LLMs were not changed or redistributed. Additionally, research followed Hugging Face's Terms of Service and Content Policy.}
    \begin{tabular}{|p{5cm}|p{1.8cm}|p{2cm}|p{3cm}|}
    \hline
    \textbf{LLM Name} & \textbf{Developer} & \textbf{Parameters} & \textbf{License} \\
    \hline
    \href{https://huggingface.co/deepseek-ai/DeepSeek-R1-Distill-Qwen-32B}{DeepSeek-R1-Distill-Qwen-32B} & DeepSeek & 32B  & MIT License\\ 
    \hline
    \href{https://huggingface.co/mistralai/Mistral-7B-Instruct-v0.3}{Mistral-7B-Instruct-v0.3} & Mistral AI & 7B & Apache 2.0 License \\
    \hline
    \href{https://huggingface.co/meta-llama/Llama-3.3-70B-Instruct}{Llama-3.3-70B-Instruct} & Meta & 70B & Llama 3.3 Community License Agreement \\
    \hline
    \href{https://huggingface.co/meta-llama/Llama-3.1-8B-Instruct}{Llama-3.1-8B-Instruct} & Meta & 8B & Llama 3.1 Community License Agreement \\
    \hline
    \href{https://huggingface.co/Qwen/Qwen2.5-72B-Instruct}{Qwen2.5-72B-Instruct} & Alibaba Cloud & 72B & Qwen License \\
    \hline
    \href{https://huggingface.co/nvidia/Llama-3.1-Nemotron-70B-Instruct-HF}{Llama-3.1-Nemotron-70B-Instruct-HF} & NVIDIA & 70B & Llama 3.1 Community License Agreement \\
    \hline
    \end{tabular}
    \label{tab:llmssummary}
\end{table}

\newpage
\subsection{Corpus of Yeast Biotechnology Paper Abstracts}
Abstracts from eight research papers were used as source material for the scientific summarisation tasks. The research papers, presented in Table \ref{tab:papers}, were open-access, sourced from arXiv, and within the domain of yeast biotechnology. The arXiv link to each paper is included within the notes section of the relevant reference in the bibliography. It should be noted that the abstracts extracted from the arXiv preprint of each paper may differ from the abstract within the published version.

\begin{table}[H]
\centering
\caption{\textbf{A List of Academic Papers Used in the Summarisation Task.} It should be noted that the abstracts extracted from the arXiv preprint of each paper may differ from the abstract within the published version.}
\begin{tabular}{|p{4cm}|p{1.3cm}|p{3cm}|p{1cm}|p{1.2cm}|p{1.2cm}|}
\hline
\textbf{Paper Title} & \textbf{Authors} & \textbf{Journal} & \textbf{Year} & \textbf{Citation} & \textbf{Abstract Length (words)} \\
\hline
Engineering Yeast Cells to Facilitate Information
Exchange & Ntetsikas, N. et al. & IEEE Transactions on Molecular, Biological, and Multi-Scale Communications & 2024 &~\cite{10429943} & 198 \\
\hline
On the Modeling of Endocytosis in Yeast & Zhang T. et al. & Biophysical Journal & 2015 &~\cite{endocytosis} & 200\\
\hline
Membrane Trafficking in the Yeast
Saccharomyces cerevisiae Model & Feyder S. et al. & International Journal of Molecular Sciences & 2015 &~\cite{membranetrafficking} & 214 \\
\hline
The biosensor based on electrochemical dynamics of fermentation in yeast
Saccharomyces cerevisiae & Kernbach S. et al. & Environmental Research & 2022 &~\cite{biosensor} & 131\\
\hline
Gateway Vectors for Efficient Artificial Gene Assembly In
Vitro and Expression in Yeast Saccharomyces cerevisiae & Giuraniuc, C.V. et al. & PLOS One & 2013 &~\cite{gateway} & 199 \\
\hline
Quantitative Analysis of the Effective Functional Structure in Yeast Glycolysis & Fuente, I.M. et al. & PLOS One & 2012 &~\cite{glycolysis} & 228 \\
\hline
The evolution of the GALactose utilization pathway in budding yeasts & Harrison, M.-C et al. & Trends in Genetics & 2022 & \cite{galactose} & 119 \\
\hline
Evolution at two levels of gene expression in yeast & Artieri, C.G. et al. & Genome Research & 2013 &~\cite{geneexpression} & 207\\
\hline
\end{tabular}
\label{tab:papers}
\end{table}

The length of each paper's abstract was collected to demonstrate that the abstract lengths are within the input token limits. The greatest abstract length was found to be 228 words, which can be approximated to around 304 tokens using the conversion estimation provided by OpenAI, where 1 token $\approx$ 0.75 words \cite{tokenconversion}.

\subsection{Prompt Methods}
\begin{figure} [ht]
    \centering
    \includegraphics[width=1\linewidth]{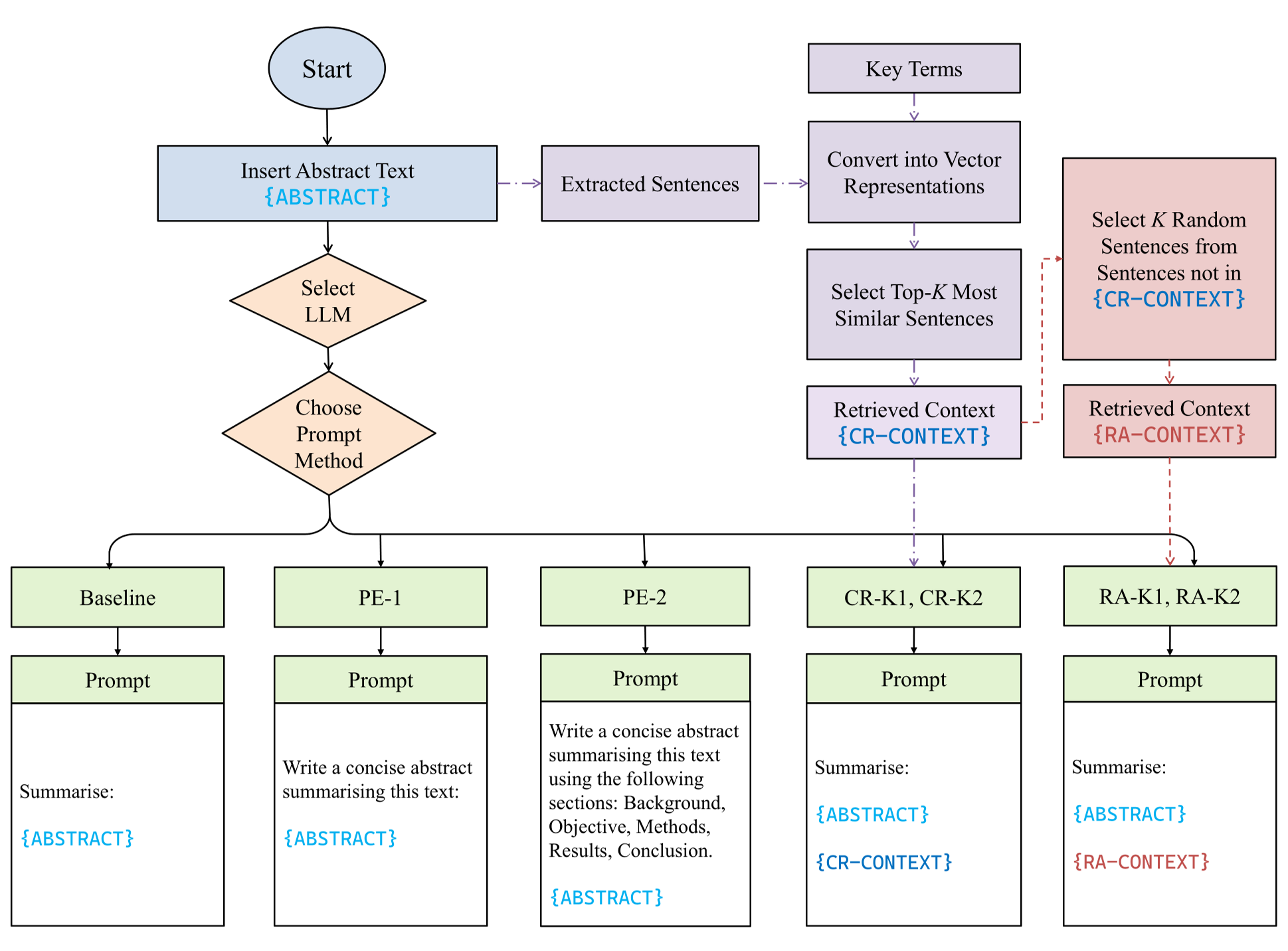}
    \caption{\textbf{Visualisation of the experimental workflow.} Each research paper abstract is inserted, and then the LLM and prompt
method are selected. Each abstract is then summarised using seven prompt methods.}
\label{fig:workflow}
\end{figure}

Figure \ref{fig:workflow} presents the workflow from the insertion of an abstract to the integration of each prompt method. Each prompt is then used to generate a summary across six LLMs, where each summary is evaluated using six metrics.

\subsubsection{Baseline Prompt}
The baseline prompt serves as a control to allow insights into the performance differences that the tested prompt approaches may cause. It was selected to be the single imperative, followed by a colon, providing the minimal amount of instruction to maximise any differences between the baseline and the engineered prompts. The colon is followed by the extracted abstract text, and implies that the user would like this text summarised, but the prompt does not explicitly request this.

\begin{quote}
\texttt{Summarise: \{ABSTRACT\}}
\end{quote}

\noindent where \texttt{\{ABSTRACT\}} represents the extracted abstract text. 

\subsubsection{Prompt Engineering Level 1 (PE-1)} 
As opposed to the baseline prompt, the first level of prompt engineering uses explicit instructions, indicating the text that the user would like to be summarised, \texttt{'summarising this text'}, and the style of summary, \texttt{'concise abstract'}. The PE-1 prompt is as follows:

\begin{quote}\small
\texttt{Write a concise abstract summarising this text: \{ABSTRACT\}}
\end{quote}

\subsubsection{Prompt Engineering Level 2 (PE-2)}
The second level of prompt engineering, PE-2, builds upon PE-1 and represents a higher complexity of instruction through the specification of the summarisation structure. The PE-2 prompt is as follows:

\begin{quote}
\texttt{Write a concise abstract summarising this text using the following sections: Background, Objective, Methods, Results, Conclusion. \{ABSTRACT\}}
\end{quote}

\subsubsection{Context Repetition (CR-K1, CR-K2)}

For $K\!\in\!\{1,2\}$, context repetition involved the identification of $K$ sentences most relevant to the prompt, where CR-K1 and CR-K2 repeat one and two key sentences, respectively. To split the abstract into sentences, the pre-trained Punkt sentence tokeniser (for English) from the Natural Language Toolkit, \texttt{NLTK 3.8.1}, data package was used. 

The key sentences from the paper abstracts were extracted using a similar but modified methodology to the top-k selection process outlined by Lui et al. (2024) \cite{liu2024datasetslargelanguagemodels} and Kukreja et al. (2024) \cite{10603291}. First, key terms were defined, consisting of terms that were considered to align with expected key abstract content. The selected key terms were \texttt{result}, \texttt{method}, and \texttt{conclusion}, and the top-$K$ sentences closest in relevance to these terms were selected as key sentences to be used within context repetition.

Each sentence was converted into a vector representation using \texttt{all-MiniLM-L6-v2}, a sentence encoder by \texttt{sentence-transformers} and hosted on Hugging Face, that is able to convert sentences into a vector within a 384-dimensional dense vector space that captures the semantics of the sentence \cite{minilm}. Similarly, vector representations of the key terms were found using \texttt{all-MiniLM-L6-v2}. To calculate the similarity between the key terms and each abstract sentence, the L2 distance between the vector representations was calculated using the \texttt{IndexFlatL2} index within the \text{FAISS}, Facebook AI Similarity Search, library. The top $K$ were selected and appended within the prompt, represented by \texttt{\{CR-CONTEXT\}}, as follows:

\begin{quote}
\texttt{Summarise: \{ABSTRACT\} \{CR-CONTEXT\}}
\end{quote} 

The prompt structure is the same as the baseline prompt to allow comparison between the baseline and CR methods. The repetition of $K$ sentences is achieved, as they are mentioned once within the abstract text, \texttt{\{ABSTRACT\}}, and a second time within the added context, \texttt{\{CR-CONTEXT\}}. $K\!\in\!\{1,2\}$ was selected for testing to evaluate whether any potential impact is influenced by the proportion that the repeated content takes up within the overall prompt. For $K=2$, the sentences consist of a larger proportion of the prompt, so any potential differences between CR-K2, RA-K2 and baseline may have higher significance.

\subsubsection{Random Addition (RA-K1, RA-K2)}
The integration of random sentences from the abstract text acts as a control to assess whether any potential effects from the context repetition experiments are due to the increased prompt length, rather than the repetition itself. Following the extraction of $K$ key sentences, the same number of random sentences were selected from the remaining sentences in the abstract. The \texttt{random.choice} method from the \texttt{NumPy} library was used to select $K$ sentences for random addition. The prompt used for RA is the same as the CR prompt, replacing \texttt{\{CR-CONTEXT\}} with \texttt{\{RA-CONTEXT\}}, which represents the random key sentences. 

\subsection{Evaluation Metrics}
The six evaluation metrics were computed using Python packages within the Hugging Face Space.
The Python \texttt{rouge} library was used to compute the ROUGE-1, ROUGE-2, and ROUGE-L values \cite{rouge-py}. The \text{bert-score} library was used to compute the BERTScore, when language was set to english, device was set to cpu, and the model type was set to \texttt{bertbert-base-uncased} \cite{bert-score-py}. The $F_1$ measure was calculated and collected. The METEOR score was accessed through the \texttt{evaluate} library within Hugging Face \cite{hfevaluate}. The cosine similarity was computed by converting the reference and summary text into vector representations using the sentence encoder \texttt{all-MiniLM-L6-v2}, and then using \texttt{cosine$\_$similarity} from the \texttt{sklearn.metrics} module within the \texttt{scikit-learn} library \cite{scikit,scikitcosine}.

The reference and summary texts are not modified prior to the computation of their overlap using these evaluation metrics, and no additional normalisation is applied. For each prompt condition, the metrics are computed using the abstract text as the reference. For the context repetition experiments using $K$ sentences, the metrics are additionally computed against $K$ key sentences.

\subsubsection{Hypothesis Mapping to Evaluation Metrics}
Hypotheses H1-H4 presented in the introduction are evaluated using the six selected evaluation metrics. H1, H2, and H4 are regarding the lexical alignment, and are therefore assessed using ROUGE-1, ROUGE-2 and ROUGE-L scores. H3 focuses on semantic alignment and is evaluated using BERTScore, METEOR score, and cosine similarity. 

\subsection{Implementation and Hugging Face Application}
In order to conduct the experiments, a custom Python application was built and deployed using Hugging Face Spaces and Gradio. The user interface was built to be interactive, consisting of a drop-down menu of LLMs, where the selected LLM would be used to generate summaries, and multiple buttons which trigger the extraction of $K$ key and random sentences, the generation of the summaries for the seven prompt conditions, and the evaluation against selected references. The summaries for each prompt condition, as well as the evaluation metrics, are displayed in the user interface once generated and computed, respectively. All the requests use the Hugging Face inference, \texttt{HF-Inference}, with the LLM-generated summaries set to a maximum token length of 300, and a timeout of 180 seconds. The sampling parameters of the LLMs were not modified, and experiments used the default configuration for each LLM. 

\section{Theory}

\subsection{Evaluation Metrics for Scientific Faithfulness}
Evaluation metrics are important for quantifying hallucinations within outputs generated by LLMs. As this research focuses on the mitigation of context inconsistency hallucinations, which are identified by LLM-response misalignment with the user input, metrics evaluating the lexical and semantic overlap between summaries and reference texts were selected. Semantic overlap measures whether two texts convey the same meaning, whereas lexical overlap measures the extent to which the words in two texts match exactly, without considering their meaning. All LLM-generated summaries were evaluated against the abstract. For the context repetition experiments, LLM-generated summaries were additionally evaluated against the repeated key sentences to determine if there was an increase in alignment with the repeated context between the baseline and context repetition methods. The evaluation metrics used within this study are ROUGE-1, ROUGE-2, ROUGE-L, BERTScore, cosine similarity, and METEOR Score.

\subsubsection{ROUGE}
ROUGE (Recall-Oriented Understudy for Gisting Evaluation),  presented by Lin (2004) \cite{lin-2004-rouge}, allows automatic summary evaluation by quantifying the lexical overlap between two texts. In contrast to precision-oriented metrics, ROUGE is recall-oriented, meaning it captures the proportion of the reference text that has been directly captured in the generated text. This allows for the evaluation of the lexical overlap between the LLM-generated summaries and the selected reference text. 

ROUGE-N is defined as:
\begin{equation}
\text{ROUGE-N} = \frac{\sum_{S \in \text{RefSummaries}} \sum_{\text{gram}_n \in S} \text{Count}_{\text{match}}(\text{gram}_n)}{\sum_{S \in \text{RefSummaries}} \sum_{\text{gram}_n \in S} \text{Count}(\text{gram}_n)}
\end{equation}

\noindent where $n$ is the length of n-gram, $(\text{gram}_n)$, $\text{Count}_{\text{match}}(\text{gram}_n)$ is the maximum number of n-grams in both the summary and reference text, and the denominator is the total number of n-grams in the reference text. Three ROUGE-N metrics are used as evaluation metrics within this work: ROUGE-1, ROUGE-2, and ROUGE-L, which respectively measure the unigram ($n=1$), bigram ($n=2$), and longest common subsequence (LCS) overlaps ($n=$LCS) between the reference and generated texts. To find the longest common subsequence between a reference text, $X$ of length $m$, and generated summary, $Y$ of length $n$, the ROUGE-L recall ($R_{LCS}$), precision ($P_{LCS}$), and F-score ($F_{LCS}$) are computed as:
\vspace{-0.4cm}

\begin{align}
\begin{array}{lll}
R_{LCS} = \dfrac{\text{LCS}(X, Y)}{m} \hspace{1cm}& 
P_{LCS} = \dfrac{\text{LCS}(X, Y)}{n} \hspace{1cm} & 
F_{LCS} = \dfrac{(1 + \beta^2) \cdot R_{LCS} \cdot P_{LCS}}{R_{LCS} + \beta^2 \cdot P_{LCS}}
\end{array}
\end{align}

\noindent where $LCS(X,Y)$ is the length of the longest common subsequence of $X$ and $Y$, and $\beta$ determines the recall and precision weightings, $\beta = P_{LCS}/R_{LCS}$. $\beta$ is commonly selected to be a large number to weight recall more heavily, making ROUGE a recall-oriented evaluation metric. As ROUGE relies on exact n-gram matches, it may incorrectly reflect cases where the summary captures the key technical points through lexically different words. Therefore, metrics that capture semantic similarity, such as BERTScore and METEOR, are used alongside ROUGE. 

\subsubsection{BERTScore}
BERTScore, presented by Zhang et al. (2020) \cite{zhang2020bertscoreevaluatingtextgeneration}, uses contextual embeddings to compare the semantic similarity of tokens between reference and generated texts. Unlike ROUGE, which depends on exact n-gram matches and lexical overlap, BERTScore depends on cosine similarity between token embeddings, hence capturing subtle semantic relationships and detecting hallucinations where the summary has diverging meaning from the reference text.

For a tokenised reference sentence, $x = \langle x_1, \ldots, x_k \rangle$, and summary sentence, $\hat{x} = \langle \hat{x}_1, \ldots, \hat{x}_l \rangle$, contextual embeddings are used to match in tokens. Match between each token in $x$ to a token in $\hat{x}$ computes recall, $R_{\text{BERT}}$, and match between each token in $\hat{x}$ to a token in $x$ computes precision, $P_{\text{BERT}}$.

\begin{equation}
    R_{\text{BERT}} = \frac{1}{|x|} \sum_{x_i \in x} \max_{\hat{x}_j \in \hat{x}} \cos(\mathbf{e}(x_i), \mathbf{e}(\hat{x}_j)), \quad
P_{\text{BERT}} = \frac{1}{|\hat{x}|} \sum_{\hat{x}_j \in \hat{x}} \max_{x_i \in x} \cos(\mathbf{e}(x_i), \mathbf{e}(\hat{x}_j))
\end{equation}

The precision and mean are combined to calculate a F1 score, which was the evaluation metric collected within this study:

\begin{equation}
    F_{\text{BERT}} = \frac{2 \cdot P_{\text{BERT}} \cdot R_{\text{BERT}}}{P_{\text{BERT}} + R_{\text{BERT}}}
\end{equation}

The F1 score was used as an evaluation metric as it combines both the precision and recall scores. Additionally, although BERTScore is able to apply inverse document frequency (IDF) weighting to emphasise domain-specific terms, this study evaluates the default BERTScore using the \texttt{bert-base-uncased} model. 

\subsubsection{Cosine Similarity}
Cosine similarity allows for semantic comparison between the embedding vectors of the LLM-generated summaries and the reference text, where embedding vectors refer to the numerical vector representations of the text.

The equation used to measure the cosine similarity between embedding vectors of the reference text, $\text{Emb}(r)$, and of the generated text, $\text{Emb}(g)$, is shown in Equation \ref{eq:cs} \cite{abdaljalil2025sindexsemanticinconsistencyindex,zhang2020bertscoreevaluatingtextgeneration}.

\begin{equation}
\text{Cosine Similarity}( \text{Emb}(r), \text{Emb}(g) ) = 
\frac{ \langle \text{Emb}(r), \text{Emb}(g) \rangle }
{ \| \text{Emb}(r) \| \cdot \| \text{Emb}(g) \| }
\label{eq:cs}
\end{equation}

Within Equation \ref{eq:cs}, $\langle \text{Emb}(r), \text{Emb}(g) \rangle$ represents the inner product of the two embedding vectors, and $\| \text{Emb}(r) \| \cdot \| \text{Emb}(g) \| $ represents the product of their Euclidean norms \cite{abdaljalil2025sindexsemanticinconsistencyindex}, the shortest distance between the origin and the embedding vectors \cite{l2norm}. Cosine similarity measures the angle between the embedding vectors and so is able to capture semantic overlap between the reference and generated texts. A cosine similarity of 1 indicates complete overlap between both embeddings \cite{abdaljalil2025sindexsemanticinconsistencyindex}. 

\subsubsection{METEOR Score}
Metric for Evaluation of Translation with Explicit Ordering (METEOR), presented by Banerjee and Lavie (2005) \cite{banerjee-lavie-2005-meteor}, calculates a score of all unigram matches between the reference and generated text, using the harmonic mean of precision and recall \cite{banerjee-lavie-2005-meteor}. METEOR Score was included as it is able to provide insights into whether the summary aligns with the extracted key technical details, even if the summary contains lexical variance from the reference text. METEOR calculates the unigram precision, $P$, recall, $R$, and harmonic mean, $F_{mean}$, as follows:

\begin{equation}
    P = \frac{m}{|s|}, \quad R = \frac{m}{|r|}, \quad F_{mean} = \frac{10 \cdot P \cdot R}{R + 9P}    
\end{equation}

\noindent where $m$ is the number of matched unigrams between the reference text, $r$, and the summary, $s$, $|s|$, is the number of unigrams in the summary, and $|r|$ is the number of unigrams in the reference text. METEOR also applies a penalty based on the number of subsequent unigram matches, referred to as $chunks$. The penalty is limited to 0.5 and calculated as follows:

\begin{equation}
    \text{Penalty} = 0.5 \cdot \left( \frac{chunks}{m} \right)
\end{equation}

\noindent where longer numbers of subsequent unigram matches result in fewer chunks, and a lower penalty. The METEOR score is then computed using $F_{mean}$ and the penalty:

\begin{equation}
    \text{METEOR} = F_{mean} \cdot (1 - \text{Penalty})
\end{equation}

\subsection{Data Analysis and Statistical Methods}
The statistical analysis intends to evaluate the four hypotheses H1-H4 presented in the introduction. H1 and H2 test whether CR and RA improve lexical alignment with the abstract. H3 tests whether PE-1 and PE-2 increase semantic alignment. H4 tests if CR increases the alignment with the repeated key sentences. 

H1-H4 involve the assessment of prompt method performance compared to the baseline prompt method, and thus can be analysed through the paired performance difference between mitigation methods. For each combination of research paper, $y \in \{1, \dots, 8\}$, LLM, $x \in \{1, \dots, 6\}$, and evaluation metric, $z \in \{1, \dots, 6\}$, where $z$= BERTScore, ROUGE-1, ROUGE-2, ROUGE-L, METEOR Score, cosine similarity, the evaluation score for each prompt method $m$, where $m$ = baseline, PE-1, PE-2, CR-K1, CR-K2, RA-K1, RA-K2 as:

\vspace{-0.4cm}
\[
S_{xyz}^{(m)}: \text{Evaluation score from metric } z \text{ for paper } y, \text{ LLM } x, \text{ under prompt method } m
\]

The paired differences in performance between each prompt method,  $m$ where $m \ne \texttt{baseline}$, and the baseline method as:

\[
\Delta_{xyz}^{(m)} = S_{xyz}^{(m)} - S_{xyz}^{(\texttt{baseline})}
\]

This allows for the calculation of the paired differences, $\Delta_{xyz}^{(m)}$, across each triplet of research paper-LLM-metric. For each set of paired differences, $\Delta_{xyz}^{(m)}$, normality was tested using the Shapiro-Wilk test \cite{shapirowilktest}. This informs the decision to conduct parametric or non-parametric significance testing. If the paired differences, $\Delta_{xyz}^{(m)}$, are found to comply with normality, a paired t-test is used as the data is organised into pairs \cite{ttest}. Otherwise, if normality assumptions are violated, a Wilcoxon signed-rank test is used \cite{wilcoxin}. In order to minimise the chance of false positives and overestimation of significance, the Bonferroni-Holm correction is applied \cite{bonferroni}. 

The significance of the paired performance difference across prompt methods is tested using two methods: (i) the Wilcoxon signed-rank test with Bonferroni-Holm correction and (ii) the bias-corrected and accelerated (BC$_a$) bootstrap method to estimate 95$\%$ confidence intervals. Only combinations that are statistically significant across both tests will be considered to support the hypothesis.

\subsubsection{Bias-corrected and Accelerated (BC$_a$) Bootstrap Method} 
\paragraph{Theory}
Within the distribution of paired differences, $\Delta_{xyz}^{(m)}$, the 95$\%$ confidence intervals (CIs) were approximated using the bias-corrected and accelerated (BC$_a$) bootstrap method. The following theory is detailed within the work by Efron (1987) \cite{efron}. The standard confidence intervals are taken using the asymptotic normal approximation, with the standard error,  $\hat{\sigma}$, taken to be a constant:

\begin{equation}
(\hat{\theta} - \theta)/\hat{\sigma} \sim \mathcal{N}(0,1)
\end{equation}

This method is first-order correct, as the term $\hat{\theta} + z_{\alpha}\hat{\sigma} $ asymptotically dominates:

\begin{equation} \label{eq:standardCI}
    \hat{\theta} + \hat{\sigma} \left( z_{\alpha} + \frac{A^{(\alpha)}}{\sqrt{n}} + \frac{B^{(\alpha)}}{n} + \cdots \right)
\end{equation}

However, as described by Efron (1987) \cite{efron}, the second-order term in Equation \ref{eq:standardCI}, $A^{(\alpha)}/\sqrt{n}$, can have a significant influence on the symmetry of confidence intervals, especially in datasets where the sample, $n$, is small. As the collected dataset within this research is considered to have a small number of research paper-LLM combinations, $n$ = 48, the use of standard intervals may cause the dataset to become biased. Therefore, the BC$_a$ bootstrap method was used. In contrast to standard intervals, the bootstrap method is second-order correct and produces the following intervals:

\begin{equation}
\hat{\theta} + \hat{\sigma} (z_{\alpha} + \frac{A^{(\alpha)}}{\sqrt{n}}  + \cdots)
\end{equation}

\noindent where asymmetry and bias are corrected with a bias-correction factor, $z_0$, and an acceleration constant,  $a$.

The BC$_a$ confidence intervals are:

\begin{equation}
\theta \in \left[ \hat{G}^{-1}(\Phi(z[\alpha])),\ \hat{G}^{-1}(\Phi(z[1 - \alpha])) \right] 
\end{equation}

\begin{equation}
z[\alpha] = z_0 + \frac{z_0 + z^{(\alpha)}}{1 - a(z_0 + z^{(\alpha)})} 
\end{equation}

The bias-correction constant,  $z_0$, is calculated as: 

\begin{equation}
z_0 = \Phi^{-1} \left( \hat{G}(\hat{\theta}) \right)
\end{equation}

The acceleration constant can be approximated as:

\begin{equation}
a \approx \frac{1}{6} \text{SKEW}_{\theta = \hat{\theta}} (i_\theta)
\end{equation}

$\hat{G}(s)$, is the cumulative distribution function of the bootstrap distribution:

\begin{equation}
\hat{G}(s) = \Pr(\hat{\theta}^* < s) = \int_{-\infty}^s f_{\theta}(\hat{\theta}^*)\, d\hat{\theta}^*
\end{equation}

\noindent $\hat{G}^{-1}$ is the inverse of $\hat{G}(s)$ and represents the percentile, which is the dataset point where $s\%$ of the bootstrapped dataset fall below that dataset point. This method is used to calculate the endpoints of the confidence interval after the corrections.

\paragraph{Application}
As the total number of research papers and LLM combinations is 48 per prompt method, the sample size is considered to be small, which may cause asymmetric confidence intervals \cite{efron}. Within the method provided by Helwig (2017) \cite{bootstrapexamples}, it was demonstrated that when no exact confidence intervals are available for a statistic, the sample median should be used. Within their provided examples, the median was selected as the statistic for $n = 50$. As the sample size per method, $n = 48$, is similar to the sample size in the example \cite{bootstrapexamples}, the median value of $\Delta_{xyz}^{(m)}$ was used as the statistic, $\hat{\theta}$. The use of the mean value may also inadvertently reflect outliers and asymmetry, so using the sample median may protect against this. Furthermore, within the bootstrapping examples \cite{bootstrapexamples}, 10000 bootstrap replicates were used, and therefore, this number was used within this research. 

\subsubsection{Hypothesis Testing}

\paragraph{BC$_a$ Bootstrap-based Hypothesis Testing}
The BC$_a$ bootstrap confidence intervals were estimated in \texttt{R} using \texttt{RStudio}, following the method outlined by Helwig (2017) \cite{bootstrapexamples}. For each prompt method, $m$, and each evaluation metric, $k$, the paired performance difference, $\Delta_{xyz}^{(m)}$, was calculated for all 48 combinations of research paper, $i$, and LLM, $j$. The median paired performance difference was used as the bootstrap statistic, $\hat{\theta}$. As per the example by Helwig (2017) \cite{bootstrapexamples}, 10000 bootstrap replicates were generated by sampling values of $\Delta_{xyz}^{(m)}$. The values were used within \texttt{boot.ci(..., type = "bca")} in \texttt{R} to calculate the 2.5th and 97.5th percentile values for each prompt method, which represent the 95$\%$ confidence interval endpoints. The following null, $H_0$ and alternative, $H_1$, hypotheses were tested for each prompt method after the calculation of the 95$\%$ confidence interval:

\begin{itemize}
  \item \textbf{H$_0$:} The 95$\%$ confidence interval of the median performance difference, $\Delta_{xyz}^{(m)}$, contains zero. This indicates that the prompt method has no change in performance when compared to the baseline prompt method.\\

  \item \textbf{H$_1$}: The 95$\%$ confidence interval of the median performance difference, $\Delta_{xyz}^{(m)}$, does not contain zero. This suggests that the prompt method does have a change in performance when compared to the baseline prompt method.
\end{itemize}

The 95$\%$ confidence interval for the median performance of the paired differences, $\Delta_{xyz}^{(m)}$, represents the range of values that the median performance could be. If the confidence interval contains a value of zero, it suggests that the median performance could have a value of zero, and the null hypothesis $H_0$ fails to be rejected. However, if the interval does not contain zero, then $H_0$ is rejected at the 5$\%$ significance level, and the median performance difference is thought to be statistically significant, indicating the tested prompt method has an impact on the summary.

\paragraph{Non-parametric Hypothesis Testing (Wilcoxon)}

The significance of the original paired differences, $\Delta_{xyz}^{(m)}$, as opposed to the bootstrapped confidence intervals, was tested using the Wilcoxon signed-rank test \cite{wilcoxin}. This test is non-parametric and assumes a non-normal distribution. While the Shapiro-Wilk test provides an indication of the normality of the dataset, non-parametric hypothesis testing will be used due to the small sample size per prompt method, $n = 48$.

The Wilcoxon test was used to test the following null hypothesis:

\begin{itemize}
  \item \textbf{H$_0$:} The median paired performance difference $\Delta_{xyz}^{(m)}$, is zero. This indicates that the prompt method has no change in performance when compared to the baseline prompt method. \\

  \item \textbf{H$_1$}: The median performance difference, $\Delta_{xyz}^{(m)}$, is not zero. This suggests that the prompt method does have a change in performance when compared to the baseline prompt method.
\end{itemize}

The collected p-values were corrected using the Bonferroni-Holm correction to minimise the overestimation of significance \cite{bonferroni}.

\section{Results and Discussion}
\subsection{Dataset Overview}
Six evaluation metrics assessing the semantic and lexical overlap between 336 LLM-generated abstract summaries and the provided abstract text were collected, resulting in a dataset of 2016 evaluation metrics. To capture further insights into the impact of context repetition on the content of LLM-generated summaries, the baseline, context repetition, and random addition prompt methods were additionally evaluated against $K$ key sentences. In total, 3744 evaluation metrics were collected across all experiments.

\subsection{Summary Statistics}

Table \ref{tab:descriptivestats} displays the mean evaluation metric scores and standard deviations across prompt method and reference text combinations, where each combination consists of 288 datapoints.

\begin{equation} \notag
\text{N = } 6\;\text{LLMs}\;\times 6\;\text{Evaluation Metrics}\;\times 8\;\text{Papers} \text{ = 288 Combinations}
\end{equation}

\vspace{-0.5cm}
\begin{equation}\notag
288\;\text{Datapoints}\;\times 13\;\text{Paper-Reference Combinations}\text{ = 3744 Total Datapoints}
\end{equation}

\begin{table} [ht]
\centering
\caption{Overview of the experimental dataset, presenting the mean score averaged across six evaluation metrics, six LLMs and eight research paper abstracts. For each combination of prompt method and reference, the standard deviation and the number of datapoints collected, N, are presented.}
\begin{tabular}{|p{3cm}|p{3cm}|p{2cm}|p{3cm}|p{1cm}|}
  \hline
\textbf{Prompt Method} &\textbf{ Reference Text} & \textbf{Mean Score} & \textbf{Standard Deviation} & \textbf{N}\\ 
  \hline
Baseline & Abstract & 0.545 & 0.224 &  288 \\
\hline
 PE-1 & Abstract & 0.505 & 0.229 &  288 \\ 
 \hline
  PE-2 & Abstract & 0.576 & 0.211 &  288 \\ 
  \hline
    CR-K1 & Abstract & 0.594 & 0.192 &  288 \\
    \hline
    RA-K1 & Abstract & 0.646 & 0.229 &  288 \\
    \hline
    CR-K2 & Abstract & 0.632 & 0.192 &  288 \\ 
    \hline
     RA-K2 & Abstract & 0.680 & 0.197 &  288 \\ 
     \hline
  Baseline & Key Sentence (K=1) & 0.263 & 0.193 &  288 \\ 
  \hline
  CR-K1 & Key Sentence (K=1) & 0.271 & 0.195 &  288 \\
  \hline
   RA-K1 & Key Sentence (K=1) & 0.311 & 0.198 &  288 \\
   \hline
  Baseline & Key Sentences (K=2) & 0.363 & 0.213 &  288 \\ 
  \hline
  CR-K2 & Key Sentences (K=2) & 0.411 & 0.207 &  288 \\
  \hline
RA-K2 & Key Sentences (K=2) & 0.502 & 0.210 &  288 \\   
   \hline
\end{tabular} \label{tab:descriptivestats}
\end{table}

\vspace*{0.2cm}

\subsection{Statistical Significance Analysis}
The dataset was first cleaned and formatted to allow for analysis using \texttt{R} within \texttt{RStudio}. A Shapiro-Wilk test was conducted to assess the normality across each research paper-LLM-metric combination. The paired performance difference, $\Delta_{xyz}^{(m)}$, between six prompt methods, $m \ne \texttt{baseline}$, and the baseline across six evaluation metrics, with the abstract text as the reference, resulted in 36 Shapiro-Wilk tests. Furthermore, $\Delta_{xyz}^{(m)}$ between the baseline and four prompt methods: CR-K1, CR-K2, RA-K1, and RA-K2, where $K$ = 1 or $K$ = 2 key sentences act as the reference, resulted in an additional 24 Shapiro-Wilk tests. Therefore, a total of 60 Shapiro-Wilk tests were performed. Of these tests, 32 combinations were normally distributed, and the remaining 28 violated normality. 

As almost half of the combinations were not normally distributed, and the dataset had a small sample size per prompt method of $n = 48$, non-parametric testing methods were selected. The paired differences were bootstrapped using the Bc$_a$ method with 10000 replicas to approximate the 95$\%$ confidence intervals for the median performance difference. To test the hypothesis in Section 4.2.1, intervals were then divided into two groups based on whether they contained zero. Moreover, Wilcoxon signed-rank tests were computed for each combination, with the Bonferroni-Holm correction applied to the collected p-values. 

The statistical significance testing demonstrated that all combinations that were found to be significant through the Wilcoxon signed-rank tests (with Bonferroni-Holm correction) were also significant across the $BC_a$ bootstrap confidence interval testing. However, 30$\%$ of the significant combinations were significant only through $BC_a$ bootstrap confidence interval testing, suggesting that the Wilcoxon signed-rank tests (with Bonferroni-Holm correction) successfully discouraged the overestimation of significance. Only combinations that were significant across both tests were selected for discussion, where the results enabled insights into significant prompt method performance effects.

\begin{figure} [H]
    \centering
    \includegraphics[width=\linewidth]{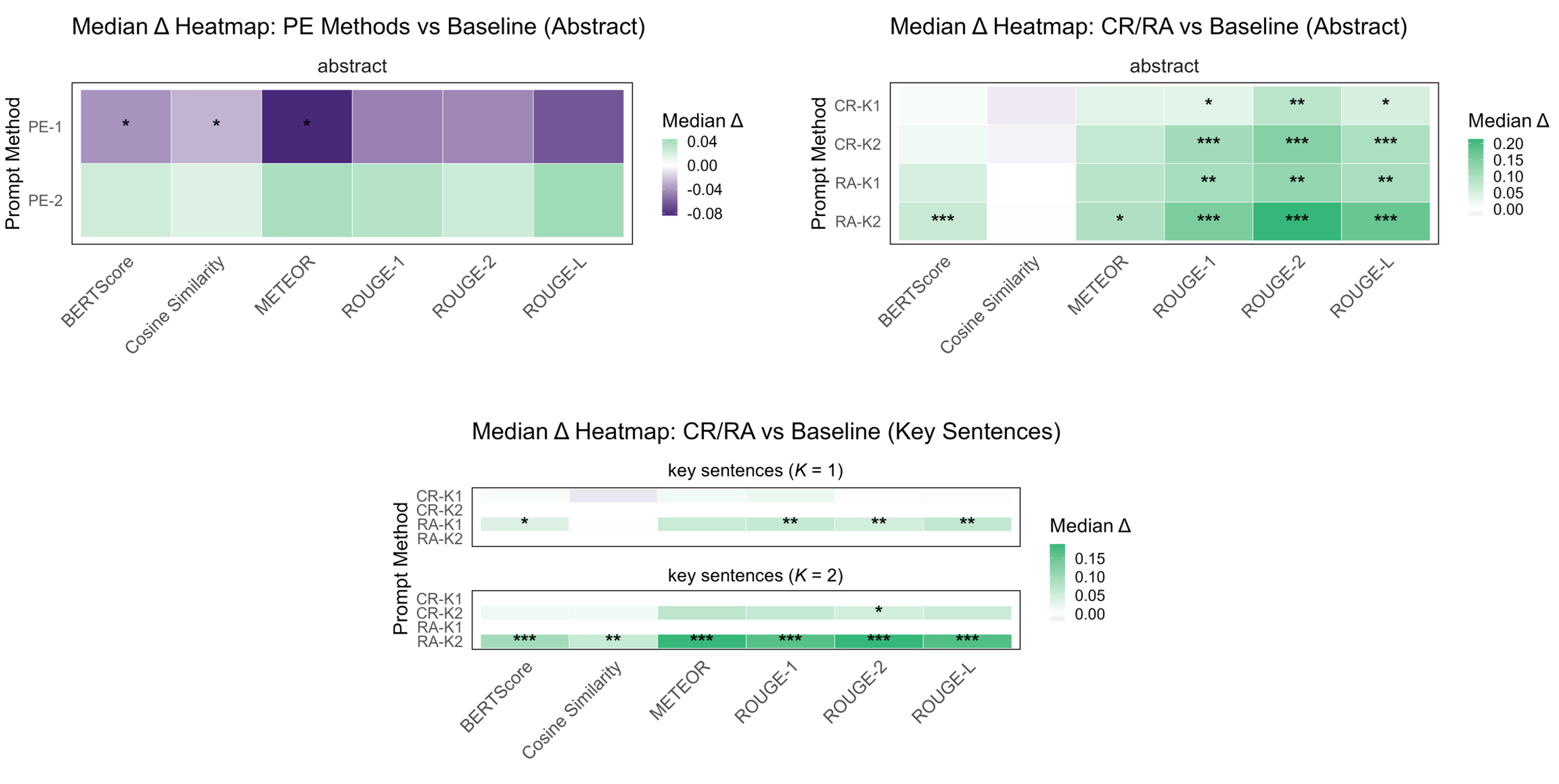}
    \caption{Median paired differences, $\Delta_{xyz}^{(m)}$, between each prompt method and the baseline across six evaluation metrics. Each cell is shaded based on the median $\Delta_{xyz}^{(m)}$ score for each combination of prompt method, evaluation metric, and reference text. Positive magnitudes are visualised in green, indicating performance improvement, and negative magnitudes are presented in purple, indicating performance decline. Asterisks identify the combinations with statistically significant performance differences, according to the BC$_a$ bootstrap confidence interval test and Bonferroni-Holm corrected Wilcoxon signed-rank test, where * = $p < 0.05$, ** = $p < 0.01$, *** = $p < 0.001$.}
    \label{fig:heatmaps}
\end{figure}

Figure \ref{fig:heatmaps} displays heatmaps illustrating the magnitude and direction of median paired differences, $\Delta_{xyz}^{(m)}$. Each cell within the heat map depicts a combination of prompt method, evaluation metric and reference type, and colour scales are used to visualise the direction and magnitude of the performance differences. Combinations that achieved statistical significance across both the corrected Wilcoxon signed rank and the BC$_a$ bootstrap confidence interval tests are identified with asterisks, where the number of asterisks indicates the extent of significance.

\begin{figure} [H]
    \centering
    \includegraphics[width=0.96\linewidth]{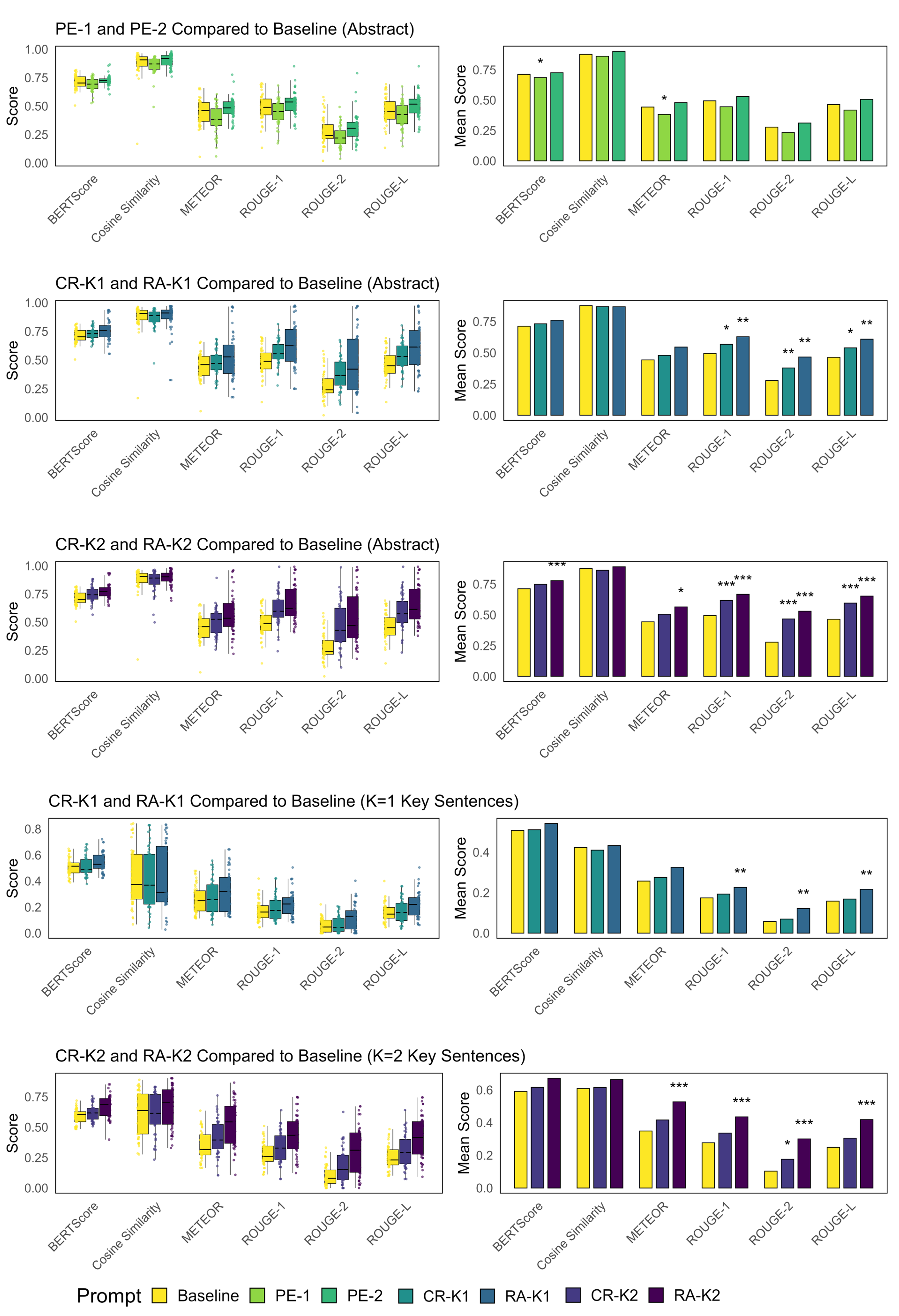}
    \caption{\textbf{Performance comparison of prompt engineering methods relative to the baseline across six evaluation metrics, evaluated against the abstract text and key sentence references}. The first row of plots visualises the impact of the first and second levels of prompt engineering methods on the alignment of the summary with the abstract text. The following four rows of plots show the comparison between the baseline, context repetition, and random addition across six evaluation metrics against the abstract text and $K$ key sentences for $K\!\in\!\{1,2\}$. Asterisks identify combinations that were significant across both the corrected Wilcoxon signed rank and the BC$_a$ bootstrap confidence interval tests.}
    \label{fig:resultboxplotsandbarcharts}
\end{figure}

Figure \ref{fig:resultboxplotsandbarcharts} compares the experimental data across each prompt method, reference type and evaluation metric. Relevant triplets were grouped and plotted together, across the six evaluation metrics, as follows: (1) \texttt{baseline}, \texttt{PE-1}, and \texttt{PE-2} against the abstract text; (2) \texttt{baseline}, \texttt{CR-K1}, and \texttt{RA-K1} against the abstract text; (3) \texttt{baseline}, \texttt{CR-K2}, and \texttt{RA-K2} against the abstract text; (4) \texttt{baseline}, \texttt{CR-K1}, and \texttt{RA-K1} against $K=1$ key sentences; and (5) \texttt{baseline}, \texttt{CR-K2}, and \texttt{RA-K2} against $K=2$ key sentences.

\subsection{Discussion}
This study intended to investigate whether prompt engineering methods can mitigate context inconsistency hallucinations in LLM summarisation of scientific abstracts. Four hypotheses H1-H4 were presented in the introduction, each focusing on the impact of different prompt methods on alignment with the reference text. H1-H4 were assessed, and the results are presented in Table \ref{tab:h1h4}.\\ 

\noindent \textit{\textbf{H1 Supported:}} Figures \ref{fig:heatmaps} and \ref{fig:resultboxplotsandbarcharts} demonstrate that context repetition had a significant positive impact on lexical alignment with the abstract text across ROUGE-1, ROUGE-2, and ROUGE-L for $K=1,2$, so H1 was supported. CR-K2 achieved higher significance than CR-K1 across all three ROUGE scores, and appears to have greater paired performance differences in Figure \ref{fig:heatmaps}, indicating a relationship between the number of key sentences repeated, $K$, and lexical alignment.\\

\noindent \textit{\textbf{H2 Supported:}} Analysis showed that random addition prompt methods significantly impacted both lexical and semantic alignment with the abstract text and repeated key sentences, so H2 was supported. For $K=2$, random addition increased semantic and lexical alignment with the repeated key sentences across all six evaluation metrics, and with the abstract across ROUGE-1, ROUGE-2, and ROUGE-L. For $K=1$, random addition improved alignment with the repeated key sentence across ROUGE-1, ROUGE-2, ROUGE-L, and BERTScore, and with the abstract across ROUGE-1, ROUGE-2, ROUGE-L, BERTScore and METEOR score. 

Additionally, the visualisation of performance effect magnitudes in Figure \ref{fig:heatmaps} reveals an asymmetrical distribution across RA-K1 and RA-K2 across both reference types. The difference between $\Delta_{xyz}^{(RA-K1)}$ and $\Delta_{xyz}^{(RA-K2)}$ is larger when the key sentences are used as a reference text, with RA-K2 achieving more positive performance effects. The asymmetry between RA-K1 and RA-K2 indicates that the repetition of two random sentences may improve lexical alignment more than the repetition of one random sentence. \\

\noindent \textit{\textbf{H3 Not Supported:}} PE-1 was found to significantly reduce semantic alignment across BERTScore, cosine similarity and METEOR score, while PE-2 performance effects were determined not to be significant. Therefore, H3 is not supported as prompt engineering did not improve semantic alignment.

Figure \ref{fig:heatmaps} illustrates the magnitude of prompt method performance effects. A small increase in prompt instruction complexity, represented by PE-1, led to a decline in semantic alignment with the abstract reference text, as evaluated by BERTScore, cosine similarity and METEOR score. This finding can be compared to related work by Kawamura et al. (2024) \cite{financialsumm}, who reported that increased prompt instruction complexity during summarisation reduced lexical alignment with the reference text, as evaluated by ROUGE-1. While both studies argue that heightened prompt instruction complexity can reduce alignment within summarisation tasks, this work identifies a decline in semantic alignment, whereas Kawamura et al. (2024) \cite{financialsumm} detect a decline in lexical alignment, and do not evaluate semantic alignment. Furthermore, the impact of elevated prompt instruction complexity on lexical alignment within summarisation was not established as significant. 

For PE-1, while significance was exhibited by semantic overlap evaluation metrics, no significant change in lexical alignment was established. The significant change in semantic overlap suggests that increased prompt instruction discouraged the LLM from paraphrasing and semantically varying summaries, while the lack of significance in lexical alignment across PE-1 and PE-2 may indicate that increased prompt instruction did not affect the exact word overlap between the reference and summary texts. Despite the lack of significance across both significance testing methods, PE-2 still achieved significance across the BC$_a$ bootstrap test across ROUGE-1 and ROUGE-L. Similarly, PE-1 delineated significance through the BC$_a$ bootstrap test within ROUGE-1, ROUGE-2, and ROUGE-L. While these results do not meet the stricter significance requirement of the corrected Wilcoxon signed-rank test, the results may propose subtle performance effects that future work could consider. \\

\noindent \textit{\textbf{H4 Partially Supported:}} The findings revealed that CR did not show a significant difference in alignment with repeated sentences for $K=1$. For $K=2$, Figures \ref{fig:heatmaps} and \ref{fig:resultboxplotsandbarcharts} indicate a significant difference in summary alignment with repeated sentences across only the ROUGE-2 evaluation scores. This indicates that H4 is weakly supported, as context repetition leads to increased lexical alignment with key sentences across ROUGE-2 for $K=2$, but not across other evaluation metrics or for $K=1$. In contrast, the random addition prompt method was found to significantly improve lexical and semantic alignment with both the key sentences and the abstract.

\begin{table}[ht]
\centering
\caption{Summary of hypothesis outcomes and interpretation.}
\begin{tabular}{|p{4cm}|p{3cm}|p{5cm}|}
\hline
\textbf{Hypothesis} & \textbf{Assessment} & \textbf{Result} \\
\hline
\textbf{H1}: Context Repetition Improves Lexical Alignment with Abstracts & Supported & CR-K1 and CR-K2 led to significant improvement in lexical alignment with the abstract compared to the baseline prompt. \\
\hline
\textbf{H2}: Random Addition Improves Lexical Alignment with Abstracts & Supported & RA-K1 and RA-K2 improved lexical alignment compared to baseline. \\
\hline
\textbf{H3}: Prompt Engineering Improves Semantic Alignment with Abstracts & Not Supported & PE-1 led to significantly reduced semantic alignment, while PE-2 showed no meaningful improvement in alignment. \\
\hline
\textbf{H4}: Context Repetition Improves Alignment with Repeated Key Sentences & Partially Supported & CR-K2 only significantly improved lexical alignment with the repeated key sentences across ROUGE-2, while CR-K1 was not significant.\\
\hline
\end{tabular}
\label{tab:h1h4}
\end{table}

Figure \ref{fig:resultboxplotsandbarcharts} presents the raw evaluation scores alongside boxplots for each prompt method and reference type. A trend in variation can be observed across the baseline, context repetition and random addition prompt methods, when the abstract serves as the reference type.

For example, RA-K2 exhibited higher variance than the CR-K2 when using the abstract as a reference, with variance percentage increases ranging from magnitudes of 4.6$\%$ for BERTScore to 41.1$\%$ and 30.9$\%$ for METEOR and cosine similarity, respectively. Despite the higher variance exhibited by RA-K2, Levene's test for homogeneity of variance found CR-K2 and RA-K2 had non-significant differences in variance (p $>$ 0.05) \cite{levenetestinR}.

Figure \ref{fig:heatmaps} allows insights into the nature of performance effects through inspection of significant evaluation metric distribution. Across the six evaluation methods, cosine similarity had the lowest rate of significance, with only around 7$\%$ of significant combinations involving cosine similarity.  In contrast,  25$\%$ of significant combinations involve ROUGE-2. Across combinations exhibiting significance within ROUGE-2, 57$\%$ were random addition prompt methods, with the remainder being context repetition prompt methods. Across all collected ROUGE-N values, the majority, 63$\%$, were random addition prompt methods, with the remainder linked to context repetition. The distribution of ROUGE-N scores, signifying lexical alignment, suggests that context repetition and random addition prompt methods disproportionately alter lexical alignment. 

The results demonstrate that some methods of prompt engineering, specifically context repetition and random addition, can significantly improve LLM summarisation performance. It was also found that increased instruction complexity can negatively impact alignment, with PE-1 even causing decreased semantic alignment.

\section{Conclusions}
This study builds upon existing work on the reduction of hallucinations in large language model (LLM) summarisation, focusing on reducing context inconsistency hallucinations. These hallucinations were quantified through the degree of semantic and lexical alignment between LLM-generated summaries and reference abstracts from yeast biotechnology research papers. Seven prompt methods were evaluated: a simple baseline prompt, two levels of prompt engineering with increasing instruction complexity (PE-1 and PE-2), two levels of context repetition using key sentences (CR-K1 and CR-K2), and two levels of random sentence addition (RA-K1 and RA-K2). Prompt method performance was evaluated across six instruction-tuned LLMs and six evaluation metrics. 

Four hypotheses were proposed in the introduction and investigated through significance testing. H1 and H2 were supported, and H3 and H4 were not supported. The results demonstrated that the prompt method can have a significant effect on the lexical and semantic alignment in zero-shot scientific summarisation. Increased instruction complexity (PE-1 and PE-2) was tested by H3 and was found not to significantly improve alignment, with PE-1 causing a decline in semantic alignment. In contrast, the random addition of abstract sentences (RA-K1 and RA-K2) achieved the highest improvement in lexical alignment, supporting H2. Context repetition (CR-K1 and CR-K2) achieved improved lexical alignment with the abstract.

This work presents prompt engineering as a practical tool for reducing context inconsistency hallucinations within LLM-based scientific summarisation. Future work could study the impact of semantic relevance of repeated sentences on summary alignment. Additionally, researching the relationships between context repetition and random addition performance effects, prompt length, and the number $K$ of key sentences could offer further insights into the utility of context repetition within LLM-based summarisation.

\begin{credits}

\subsubsection{Limitations}
While this study demonstrated that prompt engineering can be used to improve alignment in LLM summarisation, it has several limitations. The dataset size of eight abstracts limits the ability to generalise these findings across broader scientific domains without further experimentation. Additionally, automatic metrics are used within summary evaluation without human validation. Zero-shot prompting was used, indicating that the results may vary for fine-tuned LLMs. Additionally, confounding effects may be present within the context repetition and random addition prompt method evaluations, due to the varying lengths of sentences repeated. 

\subsubsection{Ethics Statement}
This research uses abstracts from publicly available scientific papers, listed in Table \ref{tab:papers} and hosted on arXiv, where the corresponding arXiv links are provided in the reference entry of each yeast biotechnology paper. Summaries were generated using LLMs under the licenses provided in Table \ref{tab:llmssummary}. All LLMs were accessed via Hugging Face inference and used according to their terms of service. The generated summaries were used solely for research purposes and were not deployed beyond the applications listed in this study. This research paper follows responsible AI practices by investigating methods to mitigate context inconsistency hallucinations, which is an important area for reducing LLM-generated misinformation within scientific summarisation.

\subsubsection{\discintname}
The authors have no competing interests to declare that are
relevant to the content of this article. 
\end{credits}

%
%
\bibliographystyle{splncs04}
\bibliography{sn-bibliography}

\end{document}